\documentclass[10pt,twocolumn,letterpaper]{article}

\usepackage{times}
\usepackage{epsfig}
\usepackage{graphicx}
\usepackage{amsmath}
\usepackage{amssymb}

\usepackage{tabularx}  
\usepackage{subfigure}  

\usepackage{xspace}

\usepackage[usenames,dvipsnames,svgnames,table]{xcolor}
\usepackage{algorithm}
\usepackage{algorithmic}
\usepackage{setspace}

\usepackage[pagebackref=true,breaklinks=true,letterpaper=true,colorlinks,bookmarks=false]{hyperref}

\begin{document}

\title{Dimensionality Reduction on Grassmannian via Riemannian Optimization:\\A Generalized Perspective}

\author{Tianci Liu$^{1,2,3}$, Zelin Shi$^{2,3}$, Yunpeng Liu$^{2,3}$\\
$^{1}$Shenyang Institute of Automation, Chinese Academy of Sciences, Shenyang 110016, China\\
$^{2}$University of Chinese Academy of Sciences, Beijing , 100049, China\\
$^{3}$Key Laboratory of Optical-Electronics Information Processing\\
}

\maketitle

\begin{abstract}
This paper proposes a generalized framework with joint normalization which learns lower-dimensional  subspaces with maximum discriminative power by making use of the Riemannian geometry. In particular, we model the similarity/dissimilarity between subspaces using various metrics defined on Grassmannian and formulate dimensionality reduction as a non-linear constraint optimization problem considering the orthogonalization. To obtain the linear mapping, we derive the components required to perform Riemannian optimization (e.g., Riemannian conjugate gradient)  from the original Grassmannian through an orthonormal projection. We respect the Riemannian geometry of the Grassmann manifold and search for this projection directly from one Grassmann manifold to another face-to-face without any additional transformations. In this natural geometry-aware way, any metric on the Grassmann manifold can be resided in our model theoretically. We have combined five metrics with our model and the learning process can be treated as an unconstrained optimization problem on a Grassmann manifold. Experiments on several datasets demonstrate that our approach leads to a significant accuracy gain over state-of-the-art methods.
\end{abstract}

\section{Introduction}

Modeling videos and image-sets by linear subspaces is shown to be  beneficial in various visual recognition tasks. However, subspaces constructed from visual data are notoriously high-dimensional, which in turn limits applicability of existing techniques. 
Consequently, the emergence of a dimensionality reduction method designed for subspaces to learn a low-dimensional and more discriminative space is extremely urgent. Furthermore, subspaces in vision have a rigorous geometry which should be concerned for the corresponding method of dimensionality reduction. Linear subspaces with the same dimensionality reside on a
special type of Riemannian manifold, i.e., Grassmann manifold, which has a nonlinear structure.
Conventional methods, such as Principal Component Analysis (PCA) \cite{holland2008principal} and Linear Discriminant Analysis (LDA) \cite{izenman2013linear}, are devised for vectors in the flat Euclidean space instead of curved Riemannian space. Simply applying these method to subspaces may occur distortions in the geometry. In this context, a natural question arises:\emph{ How can popular dimensionality reduction techniques be extended to subspaces with Riemannian geometry?}

\begin{figure}
\centering
\includegraphics[width=0.85\linewidth]{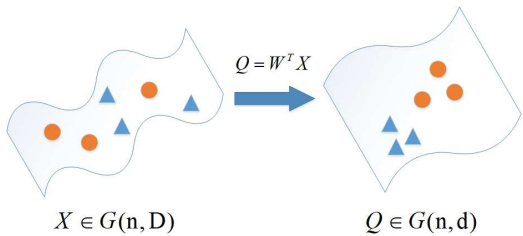}
\caption{ Conceptual illustration of the proposed supervised DR on Grassmann Manifold. The new Grassmannian leads to a lower-dimensional and more discriminative space.}
\label{shiyitu}
\end{figure}  
 
In response to this issue, this paper proposes a manifold-to-manifold method to learn a low-dimensional and more discriminative Grassmann manifold under a generalized framework, which can be regarded as a geometry-aware dimensionality reduction of the Grassmann manifold (as shown in Fig.\ref{shiyitu}). Noted that our framework is suitable for any metric on the Grassmann manifold instead of just being limited to the projection framework as \cite{huang2015projection,Wang2017Locality,Wang_2017_CVPR}. Our main contributions of this paper are: 
\begin{itemize}
\item We propose a method with the orthonormal constraint that learns a low-dimensional space of the Grassmann manifold from the high-dimensional one, which maximizes the discriminative power of the classification. 
\item  We proposed a generalized Grassmannian framework which is more extended and complete compared to other similar models \cite{huang2015projection,Wang2017Locality}. Our model is more flexible and available for various metrics on the Grassmannian. 


\item We model the dimensionality reduction as an optimization problem on Grassmannian with joint normalization considering the orthogonalization. This guarantees the reduced matrices endowed with the Grassmannian geometry in each iteration.
\end{itemize}

 The remaining parts are organized as follows: In section 2, we briefly introduce the nations of the Grassmann manifold and Riemannian metrics on it. Then the proposed method and the formulations to compute gradients are derived in section 3 and section 4. In section 5, we conduct several experiments to show the competitive performance of our approach compared with the state-of-the-art algorithms.

\section{Background}
\subsection{Grassmannian Geometry}

 A Grassmann manifold G(n,D) is the set of n-dimensional linear subspaces in the D-dimensional Euclidean space ${{\mathbb{R}}^{D}}$, which is the compact Riemannian manifold with n(D-n) dimensionality. A point on G(n,D) is a linear subspace $span(X)$, which is spanned by its orthonormal basis matrix X of size  $D \times n$ such that ${X^T}X = {I_n}$, where ${I_n}$ is the $n \times n$ identity matrix. 
In the rest of this paper, we denote X to represent its equivalence class $span(X)$ for a point on the Grassmannian.
    
On a Riemannian manifold, points are connected via smooth curves. The distance
between two points is called geodesic distance, which is defined as the length of the shortest curve connecting them on the manifold. And this shortest curve is named as geodesic.  For the Grassmannian, the geodesic distance between two points $X_1$ and $X_2$ is given by 
\begin{equation}
{d_{geo}}\left( {{X_1},{X_2}} \right) = {\left\| \Theta  \right\|_2}
\end{equation}
where $\Theta  = ({\theta _1},{\theta _2}, \cdots ,{\theta _i})$ is the vector of principal angles between  $X_1$ and $X_2$.

In addition to the geodesic distance, several other metrics related to the principal angles which approximate the true Grassmannian geodesic distance can be employed to measure the similarity between Grassmannian points.
Next,  several prevalent  Grassmannian distances which are widely used in many lectures \cite{hamm2008grassmann,edelman1998geometry,harandi2014expanding} are introduced in Tab.\ref{tab:3}.

\begin{table*}[htbp]
\centering
\caption{Different measures on the Grassmann manifold. ${X_1},{X_2}$ are two points on the Grassmannian G(n,D). }
\begin{tabular}{|c|c|c|c|}
\hline
 \textbf{Measure Name} &  \textbf{Mathematical Expression}&  \textbf{Metric/ Distance}&  \textbf{Kernel}  \\
\hline

projection F-norm \cite{edelman1998geometry}	&${d_{pro}}({X_1},{X_2}) = {2^{ - \frac{1}{2}}}{\left\| {{X_1}{X_1}^T - {X_2}{X_2}^T} \right\|_F}={\left\| {\sin (\Theta )} \right\|_2}$& ${\surd}$ &${\times}$  \\
Fuibni-Study \cite{edelman1998geometry}&$d{}_{FS}({X_1},{X_2}) = arccos\left| {\det ({X_1}^T{X_2})} \right|=\arccos (\prod\nolimits_i {\cos {\theta _i}} )$&${\surd}$&${\times}$  \\
Binet-Cauchy distance \cite{harandi2014expanding}&$d_{{\rm{BC}}}^2({X_1},{X_2}) = 2 - 2\left| {\det ({X_1}^T{X_2})} \right|=2 - 2\prod\nolimits_i {(1 - si{n^2}{\theta _i})}$&${\surd}$&${\times}$ \\
projection kernel distance \cite{harandi2014expanding}&$d_{{\rm{pk}}}^2({X_1},{X_2}) = 2{\rm{n}} - 2\left\| {{X_1}^T{X_2}} \right\|_F^2=2n - 2\sum\nolimits_i {{{\cos }^2}({\theta _i})}$&${\surd}$&${\times}$ \\
Binet-Cauchy kernel \cite{hamm2008grassmann}&$d_{{\rm{BCK}}}^2({X_1},{X_2}) = \det ({X_1}^T{X_2}{X_2}^T{X_1})=\prod\nolimits_i {\cos {}^2({\theta _i})}$&${\times}$&${\surd}$ \\

\hline
\end{tabular}
\label{tab:3}
\end{table*}

\textbf{Proposition 1.} All the metrics between two subspaces are right invariant to the group of orthonormal transformations $O(n)$. For any distance $d({X_1},{X_2})$ on the Grassmannian manifold, \[d({X_1},{X_2}) = d({X_1}{H_1},{X_2}{H_2}),{\kern 1pt} {\kern 1pt} {\kern 1pt} {\kern 1pt} {\kern 1pt} {\kern 1pt} {\kern 1pt} \forall {H_1},{H_2} \in O(n)\]

\subsection{Discriminant Learning and Dimensionality Reduction on Grassmannian}
In recent years, some related works about the Grassmannnian discriminant learning have come up. Grassmann Discriminant Analysis (GDA) \cite{hamm2008grassmann} firstly proposes a Grassmann framework which embeds the Grassmann manifold into a Reproducing Kernel Hilbert   space (RKHS) by learning a projection kernel and then learns a map to a lower-dimensional and more discriminative space under Fisher LDA criteria. Based on GDA, Graph-embedding Grassmann Discriminant Analysis (GGDA) \cite{harandi2011graph} provides a graph-embedding framework with a new Grassmannian kernel to learn a more discriminatory mapping on Grassmannian manifolds. Combining with the sparse coding and dictionary learning on Grassmann manifolds, Grassmann Dictionary Learning (GDL) \cite{harandi2013dictionary} updates a Grassmann dictionary under the projection embedding and proposes a kernelised method to solve non-linearity in the data.

However, the limitations of these methods are obvious. Firstly, the most important part is to find a desirable kernel function which is positive definite to satisfy Mercer’s theorem so that the valid RKHS can be generated. Secondly, the embedded data in the higher-dimensional Hilbert space will incur distortions through flattening the Grassmann manifold. Furthermore, the kernel functions only measure the similarity between data instead of the distance. Last but not least, the computational cost is considerably high when the size of data samples is large.

Recently, several works aim to learn the mapping from manifold to manifold directly and have attracted more and more attention. Harandi in \cite{harandi2017dimensionality} first learns a mapping with an orthonormal projection from the high-dimensional SPD manifold to a lower-dimensional and more discriminative one. 
 Projection Metric Learning on Grassmann Manifold (PML) in \cite{huang2015projection} learns a Mahalanobis-like matrix on the symmetric positive semidefinite manifold to seek a lower-dimensional and more discriminative Grassmann manifold under the projection framework by embedding Grassmann manifolds onto the space of symmetric matrices.
While \cite{huang2015projection} has reached some success, it is limited by the fixed model under the projection metric and doesn't consider about the joint normalization to maintain orthogonality during the process of optimization, which still exists a gap to the best solution for this problem. 

In this context, this paper is dedicated to a generalized Grassmannian framework to learn a lower-dimensional and more discriminative manifold without the limitation of certain metric while taking account of orthogonalization jointly during each iteration of optimization for a better solution due to respecting the geometry of Grassmannian.

\section{Joint Normalization and Dimensionality Reduction on Grassmannian}
\subsection{The proposed method}
What we want to do is to learn a mapping W, which can map the Grassmann manifold in high dimensionality G(n,D) to a lower one G(n,d) for the purpose of better classification ($D \gg d$). More specifically, supposed $X \in G(n,D)$, $Y \in G(n,d)$, we want to find a column full rank matrix $W\in {{\mathbb{R}}^{D\times d}}$ so that a  general mapping $f:G(n,D) \to G(n,d)$ can be learned:
\begin{equation}
f(X) = {W^T}X = Y
\end{equation}

This mapping can be regarded as a special kind of dimensionality reduction of the Grassmann manifold, but we aim to find a new low-dimensional geometry where two image-sets are close to each other if they belong to the same class and far apart if they don't. That is, given a set of image sets $\Gamma  = \left\{ {{X_1},{X_2}, \cdots ,{X_N}} \right\}$, this geometry is structured by the affinity matrix $G \in {R^{N \times N}}$ which is undirected and symmetric. Each element $G(i,j)$ reflects the pairwise relationship of the i-th image set and the j-th one. The detailed notation of G will be described in section~\ref{section4}. 

When the affinity matrix G is given, we plan to make use of different metrics to encode this  structure into the low-dimensional manifold. For this purpose, the objective function has the corresponding formula
\begin{equation}\label{eq3}
L(W) = \sum\limits_{i,j} {G(i,j)}  \cdot d({W^T}{X_1},{W^T}{X_2})\footnote{This is not the objective function in our model due to the lack of considering orthogonalization. We just use this one to introduce Eq.\ref{equ:4} with the joint normalization.}
\end{equation}
where $d:M \times M \to {R^ + }$ represents $d_{pro}^2$, $d{}_{FS}$, $d_{BC}^2$, $d_{BCK}^2$ or $d_{pk}^2$.

As the metrics used in Eq.\ref{eq3} are subspace distances, ${W^T}X$ should be on the Grassmannian accordingly. However, ${W^T}X$ is not guaranteed to be on the Grassmann manifold even if ${W}$ is an orthogonal matrix in general. Note that only the linear subspaces spanned by orthonormal basis matrix can form a valid Grassmann manifold. To solve this issue, we use QR-decomposition to get the orthonormal components of ${W^T}X$ s.t. 
${W^T}{X} = {Q}{R}$, where Q is the orthonormal matrix consisted of the first d columns and R is the invertible upper-triangular matrix.

\subsubsection{Joint Normalization of Y}

As shown above, considering about the orthogonalization, the cost function will be led to the following formulation:
\begin{equation}\label{equ:4}
L(W) = \sum\limits_{i,j} {G(i,j)}  \cdot d({Q_i},{Q_j})
\end{equation}
where 
\begin{equation}
{Q_i} = {W^T}{X_i}R_i^{ - 1}, {Q_j} = {W^T}{X_j}R_j^{ - 1}
\end{equation}
 In order to avoid degeneracies of optimization and follow common practice in dimensionality reduction, we impose  an orthogonality constraint on W such that ${W^T}W = {I_d}$. Finally, dimensionality reduction is written as the optimization problem which seeks the solution to W by minimizing the cost function Eq.\ref{equ:4}
\begin{equation}\label{equ:cost function}
{W^*} = \mathop {\arg \min }\limits_W L(W)\\
{\kern 1pt} {\kern 1pt} {\kern 1pt} {\kern 1pt} {\kern 1pt} {\kern 1pt} {\kern 1pt} {\kern 1pt} {\kern 1pt} {\kern 1pt} {\kern 1pt} {\kern 1pt} {\kern 1pt} {\kern 1pt} {\kern 1pt} {\kern 1pt} {\kern 1pt} {\kern 1pt} {\kern 1pt} {\kern 1pt} {\kern 1pt} {\kern 1pt} {\kern 1pt} {\kern 1pt} {\kern 1pt} {\kern 1pt} {\kern 1pt} {\kern 1pt} {\kern 1pt} {\kern 1pt} {\kern 1pt} s.t.{\kern 1pt} {\kern 1pt} {\kern 1pt} {\kern 1pt} {\kern 1pt} {\kern 1pt} {W^T}W = {I_d}
\end{equation}

From a mathematical perspective, the optimization problem with orthogonality constraints is actually an unconstrained optimization problem on the Stiefel manifold. Concretely, the search space of W is on the Stiefel manifold if the minimization problem $L(W)$ has the orthogonality constraints, i.e. ${W^T}W = {I_d}$. Moreover, if the objective function is invariant to the orthogonal group, i.e., $L(W) = L(WH)$ for any $H \in O(n)$, the search space of W is on the Grassmann manifold. 

According to Proposition \textcolor{red}{1}, it is easy to check that for any $H \in O(n)$, we have 
\begin{equation}
L(W) = L(WH)
\end{equation}
which means Eq.\ref{equ:4} is independent from the choice of basis spanned by $W$. Thus, Eq.\ref{equ:cost function} can be solved as an unconstrained minimization problem on G(d,D).

\subsection{Optimization on the Riemannian Manifold}
As mentioned above, Eq.\ref{equ:cost function} is formulated as an optimization problem on a Grassmannian. Based on recent advances in optimization on Riemannian matrix manifolds 
 \cite{absil2009optimization}, dimensionality reduction using the Grassmann or Stiefel manifolds is an emerging topic in computer vision and machine learning \cite{cunningham2015linear,hauberg2014grassmann,Wang_2017_CVPR}. In this section, a generalized framework with joint normalization of orthogonalization is developed to find a solution on the Grassmannian.

In practice, the solution to Eq.\ref{equ:cost function} can be sought through the conjugate gradient method \cite{absil2009optimization,edelman1998geometry} on G(d,D), which is implemented in the ManOpt\footnote {http://www.manopt.org} toolbox \cite{boumal2014manopt}. This nonlinear method essentially requires the gradient on the Riemannian manifold. More specifically, the Riemannian gradient on G(d,D) can be computed as
\begin{equation}
{R_W}L(W) = ({I_D} - W{W^T}){\nabla _W}L(W)
\end{equation}
where ${\nabla _W}L(W)$ is the Euclidean gradient of $L(W)$ with respect to W, which is the Jacobian matrix of size D by d.
In the next subsection, the detailed derivations of ${\nabla _W}L(W)$ under different metrics are described.

\subsection{Computing the Gradient}

\subsubsection{The Matrix Chain Rule}

In this subsection, we explain how the chain rule of matrix \cite{ionescu2015matrix} and the Taylor expansion of matrix functions \cite{magnus1999matrix} compute the partial derivative of a matrix function. For two arbitrary matrix functions, $f(X) = Y$ and $L = L(Y)$ , we have
\[L = L(Y) = L \circ f(X)\]
According to the matrix calculus theorem \cite{giles2008collected,bodewig2014matrix}, the Taylor expansions of $L(Y)$ and $L \circ f(X)$ are 
\begin{equation}\label{eq9}
L(Y + dY) - L(Y) = \frac{{\partial L}}{{\partial Y}}:dY + O\left( {{{\left\| {dY} \right\|}^2}} \right)
\end{equation}
\begin{equation}\label{eq10}
L \circ f(X + dX) - L \circ f(X) = \frac{{\partial L \circ f}}{{\partial X}}:dX + O\left( {{{\left\| {dX} \right\|}^2}} \right)
\end{equation}
where $X:Y = tr({X^T}Y)$ is an inner product for matrices.

Referred to \cite{ionescu2015matrix}, when 
\[dY = df(X;dX)\]
the Eq.\ref{eq9} and Eq.\ref{eq10} are equal. Consequently, the first order terms of the Taylor expansions in Eq.\ref{eq9} and Eq.\ref{eq10} are equal as well, which leads to the chain rule of matrix
\[\frac{{\partial L}}{{\partial Y}}:dY = \frac{{\partial L}}{{\partial Y}}:\phi (dX) \buildrel \Delta \over = {\phi ^*}(\frac{{\partial L}}{{\partial Y}}):dX = \frac{{\partial L \circ f}}{{\partial X}}:dX\]

\[{\kern 1pt} {\kern 1pt} {\kern 1pt} {\kern 1pt}  \Rightarrow \frac{{\partial L \circ f}}{{\partial X}} = {\phi ^*}(\frac{{\partial L}}{{\partial Y}})\]
where 
\[dY = \phi (dX) \buildrel \Delta \over = df(X;dX)\]
and ${\phi ^*}( \cdot )$ is a non-linear adjoint operator \cite{ionescu2015matrix} of $\phi$.

\subsubsection{The Derivation of Gradient}

Firstly, due to the joint normalization of orthogonalization during each iteration of optimization, it is necessary to calculate the partial derivatives of QR decomposition for our framework referred to \cite{Huang2016Building}.
\begin{quote}
\textbf{Proposition 2} (QR Variations). Let $X = QR$ with $X\in {{\mathbb{R}}^{m\times n}}$ and $m \ge n$, such that ${{Q}^{T}}Q=I$ with $Q\in {{\mathbb{R}}^{m\times n}}$ and R possessing the invertible upper-triangular structure with $R\in {{\mathbb{R}}^{n\times n}}$. Then
\[dQ=(I-Q{{Q}^{T}})dX{{R}^{-1}}+Q{{({{Q}^{T}}dX{{R}^{-1}})}_{atril}}\]
\[dR={{Q}^{T}}dX-{{({{Q}^{T}}dX{{R}^{-1}})}_{atril}}R\]
where ${A_{atril}} = {A_{tril}} - {({A_{tril}})^T}$, ${A_{tril}}$ denotes A with all upper-triangular elements set to 0.
Consequently the partial derivatives are
\[\begin{array}{l}
\frac{{\partial L \circ f}}{{\partial X}} = \left( {{{\left( {I - Q{Q^T}} \right)}^T}\frac{{\partial L}}{{\partial Q}} + Q{{({Q^T}\frac{{\partial L}}{{\partial Q}})}_{btril}}} \right){\left( {{R^{ - 1}}} \right)^T}\\
{\kern 1pt} {\kern 1pt} {\kern 1pt} {\kern 1pt} {\kern 1pt} {\kern 1pt} {\kern 1pt} {\kern 1pt} {\kern 1pt} {\kern 1pt} {\kern 1pt} {\kern 1pt} {\kern 1pt} {\kern 1pt} {\kern 1pt} {\kern 1pt} {\kern 1pt} {\kern 1pt} {\kern 1pt} {\kern 1pt} {\kern 1pt} {\kern 1pt} {\kern 1pt} {\kern 1pt} {\kern 1pt} {\kern 1pt} {\kern 1pt} {\kern 1pt} {\kern 1pt} {\kern 1pt} {\kern 1pt} {\kern 1pt} {\kern 1pt} {\kern 1pt} {\kern 1pt} {\kern 1pt} {\kern 1pt} {\kern 1pt} {\kern 1pt} {\kern 1pt} {\kern 1pt} {\kern 1pt} {\kern 1pt} {\kern 1pt} {\kern 1pt} {\kern 1pt} {\kern 1pt} {\kern 1pt} {\kern 1pt} {\kern 1pt}  + Q\left( {\frac{{\partial L}}{{\partial R}} - \left( {\frac{{\partial L}}{{\partial R}}{R^T}} \right){}_{btril}{{\left( {{R^{ - 1}}} \right)}^T}} \right)
\end{array}\]
where ${A_{btril}} = {A_{tril}} - {({A^T})_{tril}}$. 
\end{quote}

Next, the partial gradients of $L(W)$ in Eq.\textcolor{red}{4} with respect to W under five various metrics are computed.
Let ${X_1},{X_2} \in G(n,D)$ are two points on the Grassmannian G(n,D) and ${Y_1} = {W^T}{X_1},{Y_2} = {W^T}{X_2}$ are corresponding transformed matrices with ${{Y}_{1}},{{Y}_{2}}\in {{\mathbb{R}}^{d\times n}}$, such that ${Y_1} = {Q_1}{R_1},{Y_2} = {Q_2}{R_2}$ are obtained by QR decomposition. 
From the perspective of matrix backpropagation, the gradient computation of $L(W)$ can be split into 4 steps: 
\[W \to ({Y_1},{Y_2}) \to ({Q_1},{Q_2}) \to L\]

We consider them in reverse order from the objective down to the inputs. In the first part, the derivative of the objective function $L$ (i.e., $\frac{{\partial L}}{{\partial {Q_1}}}$, $\frac{{\partial L}}{{\partial {Q_2}}}$) can be calculated by the matrix chain rule. Then we focus on the part receiving $Y_1$ or $Y_2$ as inputs and producing the corresponding orthogonal components (i.e., $Q_1$ or $Q_2$ ). These derivatives can be obtained with the application of Proposition \textcolor{red}{2}. Finally, for the part taking $W$ as input and producing $Y_1$ or $Y_2$, the derivatives are computed by the matrix chain rule (i.e., $\frac{{\partial L \circ f}}{{\partial W}} = \frac{{\partial L}}{{\partial {Y_1}}} + \frac{{\partial L}}{{\partial {Y_2}}}$).

Subsequently, the derivatives ${\nabla _W}L(W)$ under different metrics are shown as following.

\begin{quote}
\textbf{Remark 1} (Variations under the projection metric). The resulting partial derivative of $L(W)$ is 
\end{quote}
\begin{equation}\label{equ:11}
\begin{array}{l}
\frac{{\partial L \circ f}}{{\partial W}} = {X_1}{R_1}^{ - 1}{\left( {{{\left( {I - {Q_1}{Q_1}^T} \right)}^T}\frac{{\partial L}}{{\partial {Q_1}}} + {Q_1}{{({Q_1}^T\frac{{\partial L}}{{\partial {Q_1}}})}_{btril}}} \right)^T}\\
     {\kern 1pt} {\kern 1pt} {\kern 1pt} {\kern 1pt} {\kern 1pt} {\kern 1pt} {\kern 1pt} {\kern 1pt} {\kern 1pt} {\kern 1pt} {\kern 1pt} {\kern 1pt} {\kern 1pt} {\kern 1pt} {\kern 1pt} {\kern 1pt} {\kern 1pt} {\kern 1pt} {\kern 1pt} {\kern 1pt} {\kern 1pt} {\kern 1pt} {\kern 1pt}  + {X_2}{R_2}^{ - 1}{\left( {{{\left( {I - {Q_2}{Q_2}^T} \right)}^T}\frac{{\partial L}}{{\partial {Q_2}}} + {Q_2}{{({Q_2}^T\frac{{\partial L}}{{\partial {Q_2}}})}_{btril}}} \right)^T}
\end{array}
\end{equation}
where
\[\frac{{\partial L}}{{\partial {Q_1}}} = 2\left( {{Q_1} - {Q_2}Q_2^T{Q_1}} \right)\]
\[\frac{{\partial L}}{{\partial {Q_2}}} = 2\left( {{Q_2} - {Q_1}Q_1^T{Q_2}} \right)\]

For other metrics, the form of  $\frac{{\partial L \circ f}}{{\partial W}}$ is same as Eq.\ref{equ:11} so we leave it out for the limitation of space.

\begin{quote}
\textbf{Remark 2} (Variations under the Fuibni-Study metric). The resulting partial derivative of $L(W)$ is 
\end{quote}
 \[\frac{{\partial L}}{{\partial {Q_1}}} = {Q_2}{\left( {\frac{{\partial L}}{{\partial A}}} \right)^T}, \frac{{\partial L}}{{\partial {Q_2}}} = {Q_1}\frac{{\partial L}}{{\partial A}}\]
and
 \[\frac{{\partial L}}{{\partial A}} = \frac{{ - 1}}{{\sqrt {1 - {{\left| {\det (A)} \right|}^2}} }}\left| {\det (A)} \right|{\left( {{A^{ - 1}}} \right)^T}\]
where $A = Q_1^T{Q_2}$.

\begin{quote}
\textbf{Remark 3} (Variations under the Binet-Cauchy distance). The resulting partial derivative of $L(W)$ is 
\end{quote}
 \[\frac{{\partial L}}{{\partial {Q_1}}} = {Q_2}{\left( {\frac{{\partial L}}{{\partial A}}} \right)^T}, \frac{{\partial L}}{{\partial {Q_2}}} = {Q_1}\frac{{\partial L}}{{\partial A}}\]
where 
\[\frac{{\partial L}}{{\partial A}} =  - 2\left| {\det (Q_1^T{Q_2})} \right|{\left( {Q_2^T{Q_1}} \right)^{ - 1}}\]

\begin{quote}
\textbf{Remark 4} (Variations under the projection kernel distance). The resulting partial derivative of $L(W)$ is 
\end{quote}
\[\frac{{\partial L}}{{\partial {Q_1}}} =  - 4{Q_2}Q_2^T{Q_1},\frac{{\partial L}}{{\partial {Q_2}}} =  - 4{Q_1}Q_1^T{Q_2}\]

\begin{quote}
\textbf{Remark 5} (Variations under the Binet-Cauchy kernel). The resulting partial derivative of $L(W)$ is 
\end{quote}
\[\frac{{\partial L}}{{\partial {Q_1}}} = {Q_2}Q_2^T{Q_1}\left( {\frac{{\partial L}}{{\partial A}} + {{\left( {\frac{{\partial L}}{{\partial A}}} \right)}^T}} \right)\]
\[\frac{{\partial L}}{{\partial {Q_2}}} = {Q_1}\left( {\frac{{\partial L}}{{\partial A}} + {{\left( {\frac{{\partial L}}{{\partial A}}} \right)}^T}} \right)Q_1^T{Q_2}\]
where 
\[\frac{{\partial L}}{{\partial A}} = \det (Q_1^T{Q_2}Q_2^T{Q_1}){\left( {Q_1^T{Q_2}Q_2^T{Q_1}} \right)^{ - 1}}\]

Overall, the proposed generalized framework for Grassmann manifolds with joint normalization is summarized in Algorithm \textcolor{red}{1}, where $\tau (H,{W_0},{W_1})$ denotes the parallel transport of tangent vector $H$ from ${W_0}$ to ${W_1}$.

\begin{algorithm}[htb]
\setstretch{1} 
\caption{Generalized Grassmannian DR (GGDR)}
\textbf{Input:}\\
A set of Grassmannian points $\{ {X_i}\} _{i = 1}^p$, ${X_i} \in G(n,D)$\\
The corresponding labels $\{ {y_i}\} _{i = 1}^p$, ${y_i} \in \{ 1,2, \cdots ,C\}$\\
The dimensionality $d$ of the induced manifold\\
\textbf{Output:}\\
The mapping matrix $W \in G(D,d)$\\
\\
1.Generate G using Eq.\textcolor{red}{12}, Eq.\textcolor{red}{13} and Eq.\textcolor{red}{14}\\
2.$W\leftarrow {{I}_{D\times d}}$\\
3.${W_{new}} \leftarrow W$\\
4.$H \leftarrow \textbf{0}$\\
5.\textbf{Repeat}\\
6.Normalize ${Y_i}$ according to Eq.\textcolor{red}{5} for all i.\\
7.Compute ${\nabla _W}L(W)$ by using Remark.1 to Remark.5.\\
8.${H_{new}} \leftarrow  - {R _W}L({W_{new}}) + \eta \tau (H,W,{W_{new}})$\\
9.Line search along the geodesic starting from ${W_{new}}$ in the direction ${H_{new}}$ to find ${W^*} = \arg \min L(W)$\\
10.$H \leftarrow {H_{new}}$\\
11.$W \leftarrow {W_{new}}$\\
12.${W_{new}} \leftarrow {W^*}$\\
13.\textbf{Until} convergence
\end{algorithm}

\section{Defining the Affinity Matrix}\label{section4}
 As mentioned above, the affinity matrix can be constructed according to the supervised data, which will be used in Eq.\ref{equ:4}. Noted that our framework could also apply to the unsupervised and semi-supervised settings provided that the pairwise similarity can be  measured (i.e., the measurement of distance). 
Let ${y_i}$ denote the class label of the image set ${X_i}$, with $1 \le {y_i} \le C$. Each element of the affinity matrix can be expressed as
\begin{equation}\label{equ:12}　
G(i,j) = {G_w}(i,j) - {G_b}(i,j)
\end{equation}
where ${G_w}$ is the within-class similarity graph and ${G_b}$ is the graph to measure the  between-class similarity. Eq.\ref{equ:12} resembles the Maximum Margin Criterion (MMC) of \cite{Li2006Efficient}.

${G_w}$ and ${G_b}$ are defined as
\begin{equation}
{G_w}(i,j) = \left\{ \begin{array}{l}
1,{\kern 1pt} {\kern 1pt} {\kern 1pt} {\kern 1pt} {\kern 1pt} {\kern 1pt} {\kern 1pt} if{\kern 1pt} {\kern 1pt} {\kern 1pt} {\kern 1pt} {X_i} \in {N_w}({X_j}){\kern 1pt} {\kern 1pt} {\kern 1pt} {\kern 1pt} {\kern 1pt} or{\kern 1pt} {\kern 1pt} {\kern 1pt} {\kern 1pt} {\kern 1pt} {X_j} \in {N_w}({X_i}){\kern 1pt} {\kern 1pt} \\
0,{\kern 1pt} {\kern 1pt} {\kern 1pt} {\kern 1pt} {\kern 1pt} {\kern 1pt} {\kern 1pt} {\kern 1pt} {\kern 1pt} {\kern 1pt} {\kern 1pt} {\kern 1pt} {\kern 1pt} {\kern 1pt} {\kern 1pt} {\kern 1pt} {\kern 1pt} {\kern 1pt} {\kern 1pt} {\kern 1pt} {\kern 1pt} {\kern 1pt} {\kern 1pt} {\kern 1pt} {\kern 1pt} {\kern 1pt} {\kern 1pt} {\kern 1pt} {\kern 1pt} {\kern 1pt} {\kern 1pt} {\kern 1pt} {\kern 1pt} {\kern 1pt} {\kern 1pt} {\kern 1pt} {\kern 1pt} {\kern 1pt} {\kern 1pt} {\kern 1pt} {\kern 1pt} {\kern 1pt} {\kern 1pt} {\kern 1pt} {\kern 1pt} {\kern 1pt} {\kern 1pt} {\kern 1pt} {\kern 1pt} {\kern 1pt} {\kern 1pt} {\kern 1pt} {\kern 1pt} {\kern 1pt} {\kern 1pt} {\kern 1pt} {\kern 1pt} {\kern 1pt} {\kern 1pt} {\kern 1pt} {\kern 1pt} otherwise
\end{array} \right.
\end{equation}

\begin{equation}
{G_b}(i,j) = \left\{ \begin{array}{l}
1,{\kern 1pt} {\kern 1pt} {\kern 1pt} {\kern 1pt} {\kern 1pt} {\kern 1pt} {\kern 1pt} if{\kern 1pt} {\kern 1pt} {\kern 1pt} {\kern 1pt} {X_i} \in {N_b}({X_j}){\kern 1pt} {\kern 1pt} {\kern 1pt} {\kern 1pt} {\kern 1pt} or{\kern 1pt} {\kern 1pt} {\kern 1pt} {\kern 1pt} {\kern 1pt} {X_j} \in {N_b}({X_i}){\kern 1pt} {\kern 1pt} \\
0,{\kern 1pt} {\kern 1pt} {\kern 1pt} {\kern 1pt} {\kern 1pt} {\kern 1pt} {\kern 1pt} {\kern 1pt} {\kern 1pt} {\kern 1pt} {\kern 1pt} {\kern 1pt} {\kern 1pt} {\kern 1pt} {\kern 1pt} {\kern 1pt} {\kern 1pt} {\kern 1pt} {\kern 1pt} {\kern 1pt} {\kern 1pt} {\kern 1pt} {\kern 1pt} {\kern 1pt} {\kern 1pt} {\kern 1pt} {\kern 1pt} {\kern 1pt} {\kern 1pt} {\kern 1pt} {\kern 1pt} {\kern 1pt} {\kern 1pt} {\kern 1pt} {\kern 1pt} {\kern 1pt} {\kern 1pt} {\kern 1pt} {\kern 1pt} {\kern 1pt} {\kern 1pt} {\kern 1pt} {\kern 1pt} {\kern 1pt} {\kern 1pt} {\kern 1pt} {\kern 1pt} {\kern 1pt} {\kern 1pt} {\kern 1pt} {\kern 1pt} {\kern 1pt} {\kern 1pt} {\kern 1pt} {\kern 1pt} {\kern 1pt} {\kern 1pt} {\kern 1pt} {\kern 1pt} {\kern 1pt} {\kern 1pt} otherwise
\end{array} \right.
\end{equation}
where ${N_w}({X_i}){\kern 1pt}$ consists of ${k_w}$ neighbors that belong to the same label as ${X_i}$ and ${N_b}({X_i}){\kern 1pt}$ is the set of ${k_b}$ neighbors that own different labels from  ${X_i}$. In practice, ${k_w}$ is defined as the minimum number of points in each class and the value of ${k_b} \le {k_w}$ is set by cross-validation to balance the relationship of ${G_w}$ and ${G_b}$.
\section{Experiments}
In this section, we conduct extensive experiments to evaluate our proposed method on recognition tasks. Firstly, we use the demo-data
to validate the effectiveness of our algorithm. Secondly, we evaluate our method on the Cambridge Hand Gesture dataset and the ballet dataset. Next, we conduct experiments on the Extended Yale B dataset for face recognition task. Finally, one challenging dataset for activity recognition, the JHMDB dataset, is chosen to test the performance of our method.

 In our experiments, each image set is represented in the matrix form as ${X_i} = ({x_1},{x_2},{x_3}, \cdots ,{x_n})$, where ${x_i} \in {R^D}$ corresponds to the vectorized feature of the i-th frame. 
The image set  can be constructed as a linear subspace through the singular value decomposition (SVD) of ${X_i}$.
 More specifically, we preserve the first n singular-vectors to model the linear subspace of ${X_i}$ as an element on the $G(n,D)$. In all our experiments, the dimensionality of the low-dimensional Grassmann manifold\footnote{Here, we denote the dimensionality of $G(n,d)$ as $d$ in a less rigorous way. Actually, the dimensionality of $G(n,d)$ is n(d-n).}  (i.e., $d$) and the value of $n$ are determined by cross-validation.

    To evaluate the effectiveness of our proposed algorithm, we firstly select the simple Nearest Neighbor classifier (NN) based on different Grassmannian metric. This simple classifier can clearly and directly reflect advantages of the learning lower-dimensional manifold from the original one. Next, we use three state-of-the-art algorithms for comparison, GGDA \cite{harandi2011graph}, GDL \cite{harandi2013dictionary} and PML \cite{huang2015projection}. Because both GGDA and GDL employ a kernel derived from the projection metric, we only apply them to the projection metric-based version of our method. For GGDA, the parameter $\beta $ is tuned in the range of $\left\{ {{e^1},{e^2}, \cdots ,{e^{10}}} \right\}$. For GDL, the parameter $\lambda $ is tuned in the range of $\left\{ {{e^{ - 1}},{e^{ - 2}}, \cdots ,{e^{ - 10}}} \right\}$. For PML, we use the code offered by the author and adopt the suggestions in the paper.  
Furthermore, we use GPCA, which computes the PCA of the data (i.e., ${x_i}$) in the first place, and then work with these low-dimensional descriptors. As for the popular deep learning techniques, we also concern about the VGG-net \cite{Simonyan2014Very} to obtain CNN features.
For fair comparison, the key parameters of each method are empirically tuned according to the recommendations in the original works. All algorithms used in our experiments are referred as following:
\begin{itemize}
\item NN-P/FS/PK/BC/BCK: NN classifier based on the Projection/ Fuibni-Study/ Projection kernel/ Binet-Cauchy/ Binet-Cauchy kernel metric.
\item P/FS/PK/BC/BCK-DR: NN classifier with different metric on the low-dimensional Grassmann manifold.
\item GGDA/GGDA-DR: Graph-embedding Grassmann Discriminant Analysis/ on the learning manifold.
\item GDL/GDL-DR: Grassmann Dictionary Learning/ on the learning manifold with our method.
\item  PML: Projection Metric Learning on Grassmannian.
\item  GPCA: Subspaces with low-dimensional features obtained by PCA. 
\item  VGG: Subspaces with high-dimensional CNN features obtained by VGG-net.
\end{itemize}

\begin{figure}
\centering
\includegraphics[width=\linewidth,height=4cm]{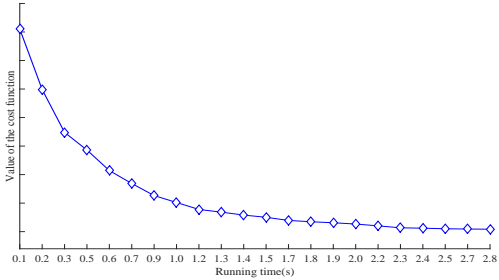}
\caption{Convergence behavior of our algorithm.}
\label{fig1}
\end{figure}
\subsection{Validation Experiment}
In this section, we choose the demo-data
 \cite{huang2015projection} to test the validity of our method. This dataset consists of 80 samples with 8 classes. Each class includes 5 training samples and 5 test samples. Each sample is a $37 \times 41$ matrix and can be represented as a point on the Grassmann manifold with the linear subspace of order 6 by SVD. In this case, we obtain 80 Grassmannian points with different labels.
Fig.\ref{fig1} illustrates the typical convergence behavior of our method. In practice, it  is observed that the algorithm generally converges speedly in less than 25 iterations. All experiments are repeated 10 times and we report the average accuracies of classification.

As can be seen, Tab.\textcolor{red}{2} shows the performance of different methods. Compared with the results of NN and DR-NN method under five metrics (i.e.,, NN-P and P-DR) , the accuracies are all improved after our mapping from the original Grassmannian to a lower-dimensional one. Especially, the result under  ${d_{FS}} $ is increased by 10 $\%$. 
Both GGDA and GDL are enhanced by our method and get the best results.
These demonstrate our method generates a better Riemannian geometry for classification (i.e., the low-dimensional Grassmann manifold). 

\begin{table*}[h]
\centering
\small
\caption{Average recognition rates on different datasets. }
\makebox[\textwidth][c]{  
\begin{tabular*}{\linewidth}{c|p{0.6cm}<{\centering}
p{0.6cm}<{\centering}p{0.6cm}<{\centering}p{0.6cm}<{\centering}p{0.6cm}<{\centering}|p{0.6cm}<{\centering}p{0.6cm}<{\centering}p{0.6cm}<{\centering}|p{0.5cm}<{\centering}p{0.55cm}<{\centering}p{0.55cm}<{\centering}p{0.55cm}<{\centering}p{0.6cm}<{\centering}p{0.6cm}<{\centering}p{1.5cm}<{\centering}p{0.6cm}<{\centering}}
\hline
 \textbf{Method} & NN-P &NN-FS	&NN-PK	&NN-BC	& NN-BCK	&GGDA	&GDL	&PML	&\textbf{P-DR} &\textbf{FS-DR}	&\textbf{PK-DR}	&\textbf{BC-DR	}&\textbf{BCK-DR}	&\textbf{GGDA-DR}	&\textbf{GDL-DR}\\
\hline\hline
\textbf{Dataset} &\multicolumn{15}{c}{Validation} \\
\hline

\textbf{Results} &  90 &\textcolor{blue}{85} &90 &85 & 85 &90 &92.5 &92.5 &\textbf{97.5} &\textcolor{blue}{95} &\textbf{97.5} &95 &92.5 &\textbf{97.5}& \textbf{97.5} \\
\hline\hline
\textbf{Dataset} &\multicolumn{15}{c}{Hand gesture} \\
\hline
\textbf{Results} & 71.11 &65.56 &67.78 &65.56 & 65.56 &\textcolor{blue}{52.22} &74.44 &55.56 &73.33 &68.89 &70.0 &66.67 &66.82 &\textcolor{blue}{73.33} &\textbf{76.67} \\
\hline\hline
\textbf{Dataset} &\multicolumn{15}{c}{Ballet} \\
\hline
\textbf{Results} & 48.63 &29.62 &48.63 &29.62 & 29.62 &37.84 &\textcolor{blue}{50.86} &51.71 &51.88 &31.42 &50.60 &31.25 &31.34 &39.38 &\textbf{\textcolor{blue}{56.68}} \\
\hline\hline
\textbf{Dataset} &\multicolumn{15}{c}{Extended Yale B} \\
\hline
\textbf{Results} & 74.21 &\textcolor{blue}{43.16} & 74.21 &43.16 & 43.16 &93.68 &95.79 &94.21 &\textbf{1} &\textcolor{blue}{92.11} &\textbf{1} &91.58 &88.95 &\textbf{1} &\textbf{1} \\
\hline
\textbf{GPCA} & 60 &38.95 & 60 &38.95 & 38.95 &86.32 & 90.53 &$- $&N/A &N/A &N/A &N/A &N/A&N/A &N/A \\
\hline\hline
\textbf{Dataset} &\multicolumn{15}{c}{JHMDB} \\
\hline
\textbf{VGG} &52.38 &\textcolor{blue}{36.51} &52.38 &36.51 & 36.51 &48.54 &52.75 &50.79 &55.56 &\textcolor{blue}{53.97} &58.73 &46.03 &42.86 &57.62 &\textbf{60.32} \\
\hline
\textbf{GPCA} & 46.03 &34.92 & 46.03 &34.92 &34.92 &45.16 &47.62 &$-$ &N/A &N/A &N/A &N/A &N/A&N/A &N/A \\
\hline\hline
\end{tabular*}
  \label{tab2}
}
\end{table*}


\subsection{Hand Gesture Recognition}
The Cambridge hand-gesture dataset \cite{kim2009canonical} contains 900 image sequences in total with 9 different classes. All sequences are divided into 5 sets according to varying illuminations. Each set consists of 180 image sequences of 10 arbitrary motions performed by 2 subjects. 
We compute Histogram of Oriented Gradient (HOG) \cite{dalal2005histograms} features to construct linear subspaces of image sequences. Our protocol is to select the first 10 sequences as test data and the last 3 sequences as training data in each class.

Tab.\ref{tab2} describes the performance of our method under different metrics and the state-of-the-art methods on this dataset. For NN method, all the metrics can reach competitive results which are more than 10$\%$ higher than PML. Specifically, P-DR is about 19$\%$ higher than PML. Both GGDA and GDL are improved on the learning lower-dimensional Grassmann manifold (i.e., GGDA-DR and GDL-DR). Furthermore, the accuracy of GGDA is boosted in a large extent by more than 21$\%$ (i.e., from 52.22$\%$ to 73.33$\%$). GDL-DR improves the original method (i.e. GDL) and reaches the best performance. 



\subsection{Recognition on the Ballet Dataset}
 The ballet dataset includes 440 videos which are derived from an instructional ballet DVD \cite{wang2009human}. All these videos can be classified into 8 complicated motion patterns acted by 3 persons. This dataset is challenging because the large intra-class variations exist in respects of spatial and temporal scale, clothing, speed and movement.

We generate 1328 image sets from the dataset by grouping every 12 frames which derive from the same action into one image set. Each image set is represented as a subspace based on the HOG feature. We select 20 image sets from each action (i.e., 160 samples in total) as training data and 1168 samples for test. For the projection metric and projection kernel metric, each image set is represented as a linear subspace of order 6. For other metrics, the dimension of each subspace is set to 3.

Tab.\ref{tab2} shows the experimental evaluation on this dataset. From the results, we can observe that accuracies of NN classifier on the learning Grassmann manifold are always improved compared with results on the original manifold. P-DR not only outperforms GGDA about 15$\%$, but also outperforms PML and GDL. For the best result of GDL-DR, our method boosts the accuracy of GDL on the learning Grassmann manifold  about 6$\%$ and get 56.68$\%$.

\subsection{Face Recognition}
The Extended Yale B dataset \cite{KCLee05} contains 2432 face images of 38 human subjects under 64 different illumination conditions.  All images used are front face images which are manually aligned, cropped, and then resized to the size 20x20.  We construct each image set to form a subspace matrix by randomly choosing 4 images from each individual. We totally generate 380 Grassmann points from 38 subjects, which means that 10 image sets are produced for each individual. Finally, we choose 190 samples for training and the rest for testing.

The results are shown in Tab.\ref{tab2}. The accuracies on the new Grassmannian under  ${d_{pro}} $ and  ${d_{pk}} $ (i.e., P-DR and PK-DR ) are enhanced by 25$\%$ to the best result. Although the results on G(n,d) under other three metrics are around 90$\%$, they are improved by a large extent about 45$\%$, which catches up with the state-of-the-arts. Both the GDL-DR and the GGDA-DR improve the original methods and outperforms the other competing methods.

\subsection{Action Recognition on the JHMDB Dataset}
JHMDB Dataset \cite{Jhuang2013Towards} is a challenging dataset on activity recognition due to the  complex and changeable scenarios. It contains more variations on scenes and
viewpoints, which can be used to examine the robustness of the proposed methods in noised scenarios.
We employ the Matconvnet \cite{Vedaldi2015MatConvNet} to extract the state-of-the-art CNN features on the FC6\footnote{We use the rectified output of the fully-connected layer fc6 of the VGG-net for all our evaluations which are 4096 dimensional vectors.} layer of  the standard 16-layer VGG-net deep learning network \cite{Simonyan2014Very} (i.e., VGG-16 Net). The VGG-16 Net model is pre-trained on Imagenet \cite{Deng2009ImageNet}, and then fine-tuned on the data from the training sets of JHMDB and UCF101 \cite{Soomro2012UCF101} datasets. We generate 2730 samples (i.e., 1260 test samples and 1470 training samples) of G(10,4096) and report the performances in Tab.\textcolor{red}{2}. In the original manifold, NN-P and NN-PK are better than other three metrics in a large margin. However, after the dimensionality reduction, the results under five metrics reach a competitive level enhanced by our method. Based on NN classifier, PK-DR arrives the best performance with a promotion of more than 6$\%$ and NN-FS has a
significant gain of around 17$\%$ especially. For the state-of-the-art methods,  the results of GGDA and GDL are raised by about 9$\%$ and 8$\%$ respectively.

\begin{figure}
\centering
\includegraphics[width=\linewidth,height=4cm]{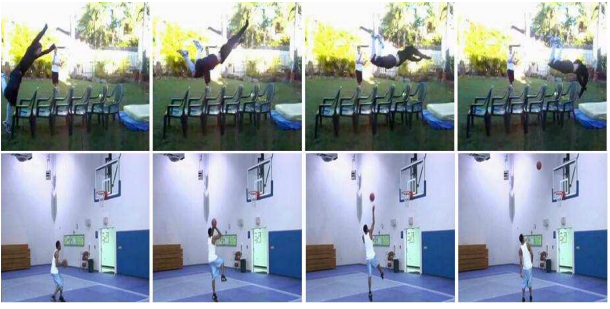}
\caption{Examples of the JHMDB dataset.}
\end{figure}

\section{Discussion}
The general convergence analysis of Newton method on Grassmannian has described theoretically in e.g. \cite{absil2004riemannian}. Our method also provides a desirable convergence  performance as shown in Fig.\ref{fig1}. Thanks to the Riemannian geometry obtained by joint normalization, the optimization always iterates towards a good solution with the mild influence of the initialization W.

\begin{figure}
\subfigure[]{
\begin{minipage}[t]{0.2\textwidth}
\includegraphics[width=4.6cm,height=4.6cm]{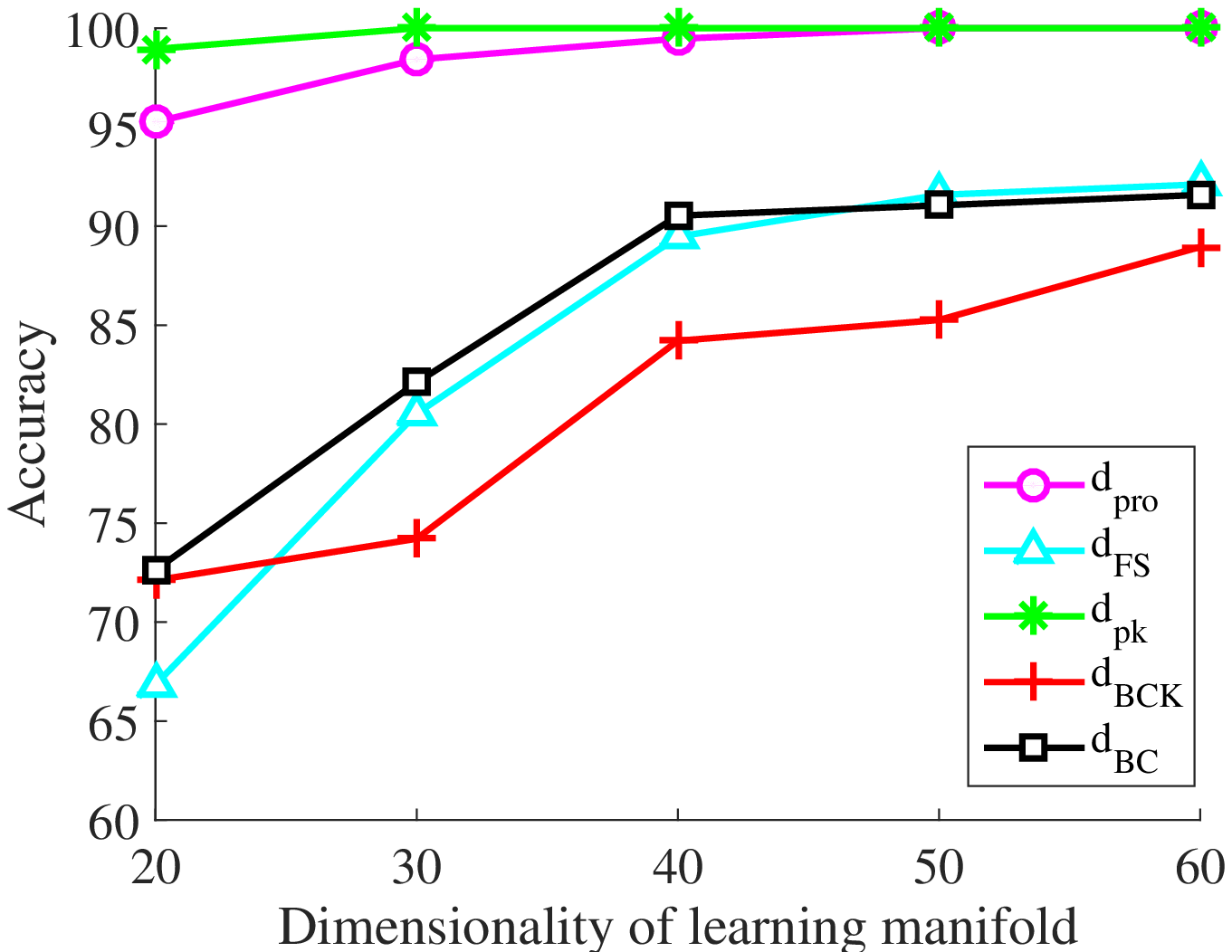} 
\end{minipage}
}
\subfigure[]{
\begin{minipage}[t]{0.3\textwidth}
\flushright
\includegraphics[width=4.7cm,height=4.7cm]{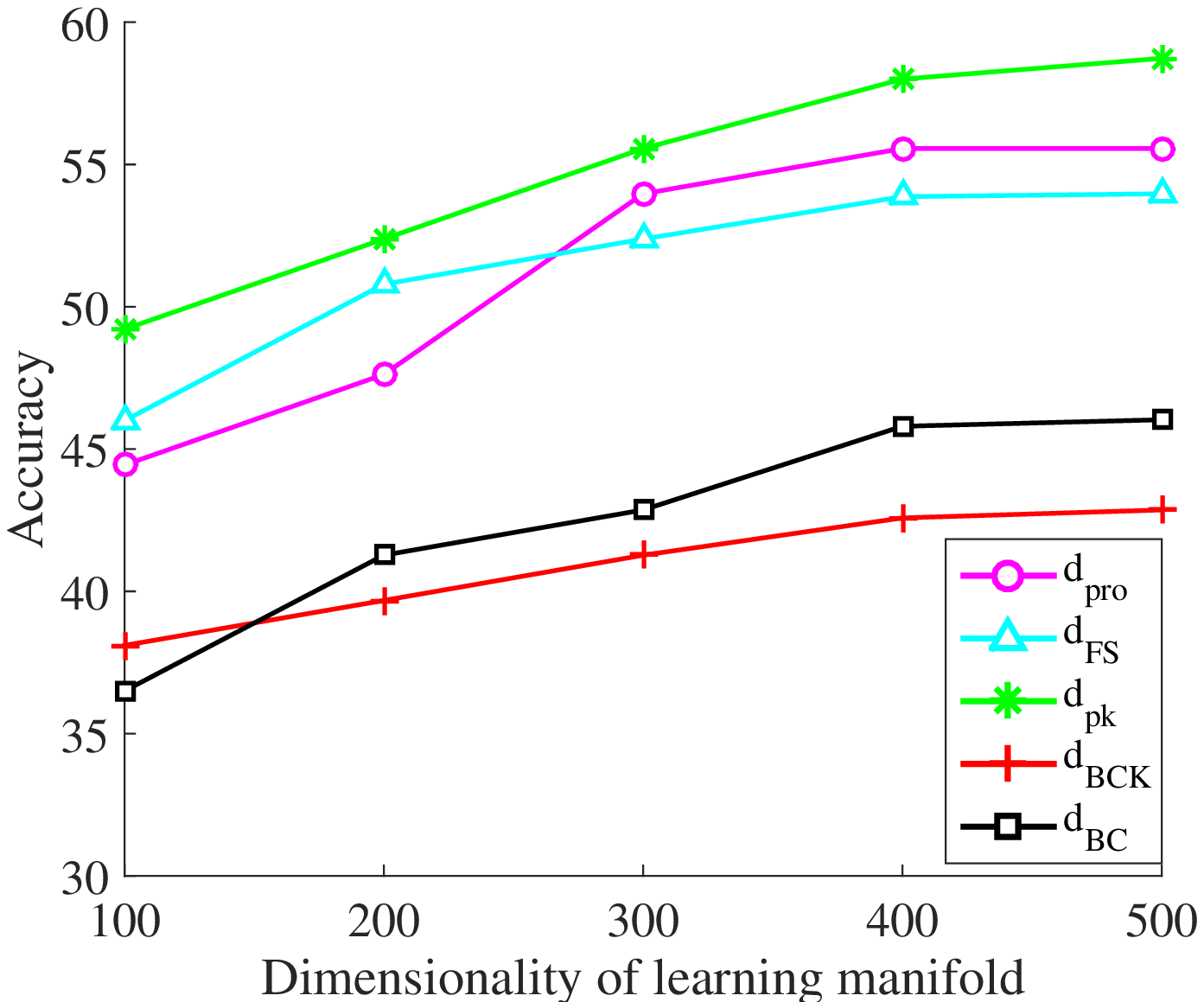} 
\end{minipage}
}
\caption{Averaged accuracies of the proposed method with different
dimensionalities of the learning manifold: (a) The curves of five metrics for the Extended Yale B dataset. (b) The curves of five metrics for JHMDB dataset.}
\label{fig3}
\end{figure}

Because our paper aims to learn a discriminative low-dimensional manifold $G(n,d)$ from $G(n,D)$,  the impact of the setting of the reduced dimensionality (i.e., $d$) should also be cared about. For this reason, the performances of our method with different $d$ on the Extended Yale B and JHMDB datasets are reported in Fig.\ref{fig3}. It is gratifying that our method gets a pleased accuracy with a desirable low dimensionality and the impact of $d$ tends to be mild when it is large enough.

Finally, we compare the capabilities of the five metrics in Table 1. From the experimental evaluations above, ${d_{pro}} $ and  ${d_{pk}} $ obtain better results than remaining metrics(i.e., ${d_{FS}} $, ${d_{BC}} $ and ${d_{BCK}} $). This maybe occurs by the reason that these metrics compute the distances related to $cosine$ or $sine$ of the principle angles between subspaces. ${d_{pro}} $ and  ${d_{pk}}$ are based on the accumulation operation while the others are based on multiplications. Luckily, these metrics lead to competitive results on the learning manifold by our method. We also compare the running time of our method under different metrics on the Extended Yale B dataset. The time cost of ${d_{pro}} $ is high because the computation complex of ${X_i}{X_j^T}$ is more expensive than that of ${X_i^T}{X_j}$ when the subspace dimension is large. From the perspectives of the time cost and accuracy, ${d_{pk}}$ is a good choice for our model.
\begin{table}[htbp]
\centering
\small
\begin{tabular}{|c|c|c|c|c|c|}
\hline
 \textbf{Metric} &  \textbf{${d_{pro}} $}&  \textbf{${d_{FS}} $}&  \textbf{${d_{pk}} $} &  \textbf{${d_{BC}} $}&  \textbf{${d_{BCK}} $} \\
\hline
 \textbf{NN} &  88.26 & 0.77& 0.89 &  0.78&  1.12 \\
\hline
 \textbf{DR-NN} & 1.71& 0.50& 0.64 &  0.49 &  0.58 \\
\hline
\end{tabular}
\caption{Running time (seconds) under different metrics. }
\end{table}

\section{Conclusion and Future Work}
To the best of our knowledge, this work is the first attempt to provide a generalized Grassmannian framework without certain metric limitation and shows the importance of respecting the Riemannian geometry when performing dimensionality reduction with joint normalization for orthogonality. 

We introduced a novel supervised algorithm inherently
to learn a low-dimensional and more discriminative Grassmann manifold from the original one under different metrics. 
Learning can be modeled as an optimization problem on a Grassmann manifold. Our new approach can not only serve as a dimensionality reduction method but also a discriminant learning technique for the Grassmann manifold. Our experimental evaluation has demonstrated that the resulting low-dimensional Grassmann manifold  leads to state-of-the-art recognition accuracies on several challenging datasets.
In the future, we plan to study more cost functions and metrics within our framework for more discriminative capability. Furthermore, we intend to extend our method to unsupervised and semi-supervised scenarios.

\

{\small
\bibliographystyle{ieee}
\bibliography{sample}

\begin{thebibliography}{10}\itemsep=-1pt

\bibitem{absil2004riemannian}
P.-A. Absil, R.~Mahony, and R.~Sepulchre.
\newblock Riemannian geometry of grassmann manifolds with a view on algorithmic
  computation.
\newblock {\em Acta Applicandae Mathematicae}, 80(2):199--220, 2004.

\bibitem{absil2009optimization}
P.-A. Absil, R.~Mahony, and R.~Sepulchre.
\newblock {\em Optimization algorithms on matrix manifolds}.
\newblock Princeton University Press, 2009.

\bibitem{bodewig2014matrix}
E.~Bodewig.
\newblock {\em Matrix calculus}.
\newblock Elsevier, 2014.

\bibitem{boumal2014manopt}
N.~Boumal, B.~Mishra, P.-A. Absil, R.~Sepulchre, et~al.
\newblock Manopt, a matlab toolbox for optimization on manifolds.
\newblock {\em Journal of Machine Learning Research}, 15(1):1455--1459, 2014.

\bibitem{cunningham2015linear}
J.~P. Cunningham and Z.~Ghahramani.
\newblock Linear dimensionality reduction: survey, insights, and
  generalizations.
\newblock {\em Journal of Machine Learning Research}, 16(1):2859--2900, 2015.

\bibitem{dalal2005histograms}
N.~Dalal and B.~Triggs.
\newblock Histograms of oriented gradients for human detection.
\newblock In {\em Computer Vision and Pattern Recognition, 2005. CVPR 2005.
  IEEE Computer Society Conference on}, volume~1, pages 886--893. IEEE, 2005.

\bibitem{Deng2009ImageNet}
J.~Deng, W.~Dong, R.~Socher, L.~J. Li, K.~Li, and F.~F. Li.
\newblock Imagenet: A large-scale hierarchical image database.
\newblock In {\em Computer Vision and Pattern Recognition, 2009. CVPR 2009.
  IEEE Conference on}, pages 248--255, 2009.

\bibitem{edelman1998geometry}
A.~Edelman, T.~A. Arias, and S.~T. Smith.
\newblock The geometry of algorithms with orthogonality constraints.
\newblock {\em SIAM journal on Matrix Analysis and Applications},
  20(2):303--353, 1998.

\bibitem{giles2008collected}
M.~B. Giles.
\newblock Collected matrix derivative results for forward and reverse mode
  algorithmic differentiation.
\newblock In {\em Advances in Automatic Differentiation}, pages 35--44.
  Springer, 2008.

\bibitem{hamm2008grassmann}
J.~Hamm and D.~D. Lee.
\newblock Grassmann discriminant analysis: a unifying view on subspace-based
  learning.
\newblock In {\em Proceedings of the 25th international conference on Machine
  learning}, pages 376--383. ACM, 2008.

\bibitem{harandi2017dimensionality}
M.~Harandi, M.~Salzmann, and R.~Hartley.
\newblock Dimensionality reduction on spd manifolds: The emergence of
  geometry-aware methods.
\newblock {\em IEEE transactions on pattern analysis and machine intelligence},
  2017.

\bibitem{harandi2013dictionary}
M.~Harandi, C.~Sanderson, C.~Shen, and B.~C. Lovell.
\newblock Dictionary learning and sparse coding on grassmann manifolds: An
  extrinsic solution.
\newblock In {\em Proceedings of the IEEE International Conference on Computer
  Vision}, pages 3120--3127, 2013.

\bibitem{harandi2014expanding}
M.~T. Harandi, M.~Salzmann, S.~Jayasumana, R.~Hartley, and H.~Li.
\newblock Expanding the family of grassmannian kernels: An embedding
  perspective.
\newblock In {\em European Conference on Computer Vision}, pages 408--423.
  Springer, 2014.

\bibitem{harandi2011graph}
M.~T. Harandi, C.~Sanderson, S.~Shirazi, and B.~C. Lovell.
\newblock Graph embedding discriminant analysis on grassmannian manifolds for
  improved image set matching.
\newblock In {\em Computer Vision and Pattern Recognition (CVPR), 2011 IEEE
  Conference on}, pages 2705--2712. IEEE, 2011.

\bibitem{hauberg2014grassmann}
S.~Hauberg, A.~Feragen, and M.~J. Black.
\newblock Grassmann averages for scalable robust pca.
\newblock In {\em Proceedings of the IEEE Conference on Computer Vision and
  Pattern Recognition}, pages 3810--3817, 2014.

\bibitem{holland2008principal}
S.~M. Holland.
\newblock Principal components analysis (pca).
\newblock {\em Department of Geology, University of Georgia, Athens, GA}, pages
  30602--2501, 2008.

\bibitem{huang2015projection}
Z.~Huang, R.~Wang, S.~Shan, and X.~Chen.
\newblock Projection metric learning on grassmann manifold with application to
  video based face recognition.
\newblock In {\em Proceedings of the IEEE Conference on Computer Vision and
  Pattern Recognition}, pages 140--149, 2015.

\bibitem{Huang2016Building}
Z.~Huang, J.~Wu, and L.~Van~Gool.
\newblock Building deep networks on grassmann manifolds.
\newblock 2016.

\bibitem{ionescu2015matrix}
C.~Ionescu, O.~Vantzos, and C.~Sminchisescu.
\newblock Matrix backpropagation for deep networks with structured layers.
\newblock In {\em Proceedings of the IEEE International Conference on Computer
  Vision}, pages 2965--2973, 2015.

\bibitem{izenman2013linear}
A.~J. Izenman.
\newblock Linear discriminant analysis.
\newblock In {\em Modern multivariate statistical techniques}, pages 237--280.
  Springer, 2013.

\bibitem{Jhuang2013Towards}
H.~Jhuang, J.~Gall, S.~Zuffi, C.~Schmid, and M.~J. Black.
\newblock Towards understanding action recognition.
\newblock In {\em IEEE International Conference on Computer Vision}, pages
  3192--3199, 2013.

\bibitem{kim2009canonical}
T.-K. Kim and R.~Cipolla.
\newblock Canonical correlation analysis of video volume tensors for action
  categorization and detection.
\newblock {\em IEEE Transactions on Pattern Analysis and Machine Intelligence},
  31(8):1415--1428, 2009.

\bibitem{KCLee05}
K.~Lee, J.~Ho, and D.~Kriegman.
\newblock Acquiring linear subspaces for face recognition under variable
  lighting.
\newblock {\em IEEE Trans. Pattern Anal. Mach. Intelligence}, 27(5):684--698,
  2005.

\bibitem{Li2006Efficient}
X.~R. Li, T.~Jiang, and K.~Zhang.
\newblock Efficient and robust feature extraction by maximum margin criterion.
\newblock {\em IEEE Transactions on Neural Networks}, 17(1):157--165, 2006.

\bibitem{magnus1999matrix}
J.~R. Magnus and H.~Neudecker.
\newblock Matrix differential calculus with applications in statistics and
  econometrics.
\newblock 1999.

\bibitem{Simonyan2014Very}
K.~Simonyan and A.~Zisserman.
\newblock Very deep convolutional networks for large-scale image recognition.
\newblock {\em Computer Science}, 2014.

\bibitem{Soomro2012UCF101}
K.~Soomro, A.~R. Zamir, and M.~Shah.
\newblock Ucf101: A dataset of 101 human actions classes from videos in the
  wild.
\newblock {\em Computer Science}, 2012.

\bibitem{Vedaldi2015MatConvNet}
A.~Vedaldi and K.~Lenc.
\newblock Matconvnet: Convolutional neural networks for matlab.
\newblock In {\em ACM International Conference on Multimedia}, pages 689--692,
  2015.

\bibitem{Wang2017Locality}
B.~Wang, Y.~Hu, J.~Gao, Y.~Sun, H.~Chen, M.~Ali, and B.~Yin.
\newblock Locality preserving projections for grassmann manifold.
\newblock In {\em Twenty-Sixth International Joint Conference on Artificial
  Intelligence}, pages 2893--2900, 2017.

\bibitem{Wang_2017_CVPR}
Q.~Wang, J.~Gao, and H.~Li.
\newblock Grassmannian manifold optimization assisted sparse spectral
  clustering.
\newblock In {\em The IEEE Conference on Computer Vision and Pattern
  Recognition (CVPR)}, July 2017.

\bibitem{wang2009human}
Y.~Wang and G.~Mori.
\newblock Human action recognition by semilatent topic models.
\newblock {\em IEEE transactions on pattern analysis and machine intelligence},
  31(10):1762--1774, 2009.

\end{thebibliography}
}

\end{document}